\documentclass{article}
\usepackage{ijcai17}

\usepackage{times}
\usepackage{epsfig}
\usepackage{graphicx}
\usepackage{amsmath}
\usepackage{amssymb}
\usepackage{bm}
\usepackage{algorithm}
\usepackage{algpseudocode}
\usepackage[tight,footnotesize]{subfigure}

\title{Salient Object Detection with Semantic Priors}

\author{Tam V. Nguyen\\
	Department of Computer Science \\ University of Dayton \\tamnguyen@udayton.edu
	\And
	Luoqi Liu \\ Department of ECE \\ National University of Singapore\\liuluoqi@u.nus.edu }

\begin{document}

\maketitle

\begin{abstract}
Salient object detection has increasingly become a popular topic in cognitive and computational sciences, including computer vision and artificial intelligence research. In this paper, we propose integrating \textit{semantic priors} into the salient object detection process. Our algorithm consists of three basic steps. Firstly, the explicit saliency map is obtained based on the semantic segmentation refined by the explicit saliency priors learned from the data. Next, the implicit saliency map is computed based on a trained model which maps the implicit saliency priors embedded into regional features with the saliency values. Finally, the explicit semantic map and the implicit map are adaptively fused to form a pixel-accurate saliency map which uniformly covers the objects of interest. We further evaluate the proposed framework on two challenging datasets, namely, ECSSD and HKUIS. The extensive experimental results demonstrate that our method outperforms other state-of-the-art methods.
\end{abstract}

\section{Introduction}
Salient object detection aims to determine the salient objects which draw the attention of humans on the input image. It has been successfully adopted in many practical scenarios, including image resizing~\cite{CA}, attention retargeting~\cite{reattention}, dynamic captioning~\cite{ACMMM13} and video classification~\cite{STAP}. The existing methods can be classified into biologically-inspired and learning-based approaches.

The early \textbf{biologically-inspired} approaches \cite{IT,Koch} focused on the contrast of low-level features such as orientation of edges, or direction of movement. Since human vision is sensitive to color, different approaches use local or global analysis of (color-) contrast. Local methods estimate the saliency of a particular image region based on immediate image neighborhoods, e.g., based on histogram analysis~\cite{RC}. While such approaches are able to produce less blurry saliency maps, they are agnostic of global relations and structures, and they may also be more sensitive to high frequency content like image edges and noise. In a global manner, \cite{FT} achieves globally consistent results by computing color dissimilarities to the mean image color. There also exist various patch-based methods which estimate dissimilarity between image patches \cite{CA,SF}. While these algorithms are more consistent in terms of global image structures, they suffer from the involved combinatorial complexity, hence they are applicable only to relatively low resolution images, or they need to operate in spaces of reduced image dimensionality \cite{AIM}, resulting in loss of salient details and highlighting edges.

\begin{figure*}[!t]
\centering
\includegraphics[width = \linewidth]{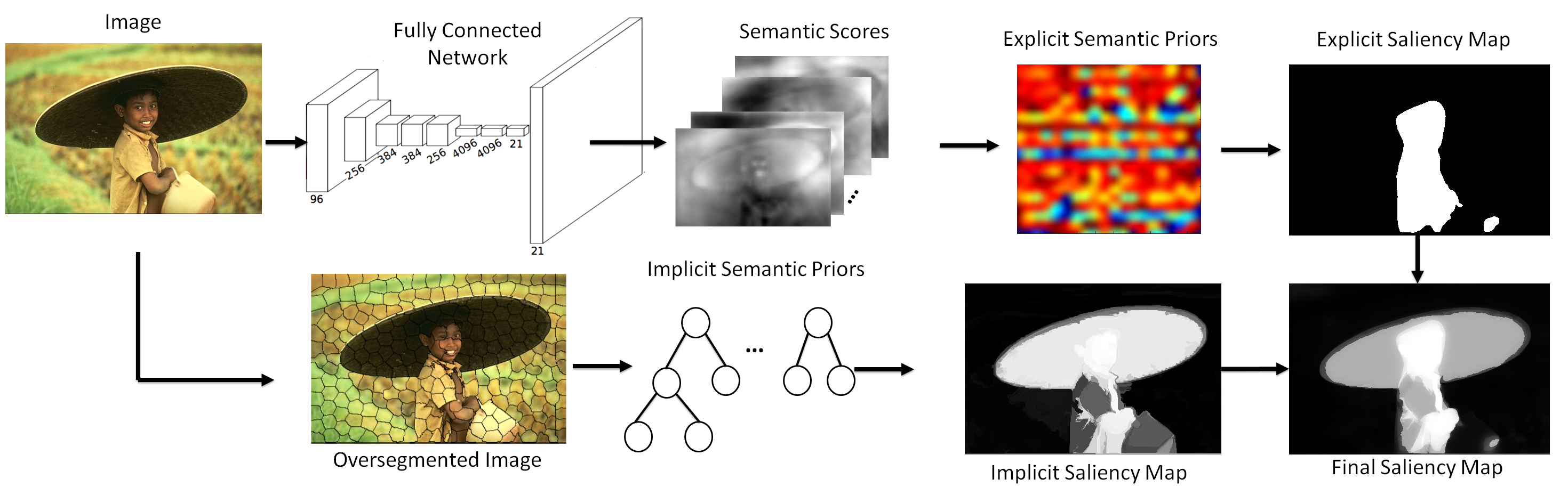}
\caption{Pipeline of our SP saliency detection algorithm: semantic scores from the semantic extraction (Section~\ref{sec:21}), explicit semantic priors to compute the explicit map (Section~\ref{sec:22}), implicit semantic priors to compute the implicit map (Section~\ref{sec:23}), and saliency fusion (Section~\ref{sec:24}).}
\label{fig:framework}
\end{figure*}

For the \textbf{learning-based} approaches,~\cite{DRFI} trained a model to learn the mapping between regional features and saliency values. Meanwhile,~\cite{HDCT} separated the salient regions from the background by finding an optimal linear combination of color coefficients in the high-dimensional color space. However, the resulting saliency maps tend to also highlight adjacent regions of salient object(s). Additionally, there exist many efforts to study visual saliency with different cues, \textit{i.e.}, depth matters~\cite{Depth}, audio source~\cite{Audio}, touch behavior~\cite{Touch}, and object proposals~\cite{AH}. 

Along with the advancements in the field, a new challenging question is arisen ``why an object is more salient than others''. This emerging question appears along with the rapid evolvement of the research field. The early datasets, \textit{i.e.}, MSRA1000~\cite{FT}, only contain images with one single object and simple background. The challenge is getting more serious when more complicated saliency datasets, ECSSD~\cite{HS} and HKUIS~\cite{HKUIS} are introduced with one or multiple objects in an image with complex background. This drives us to the difference between the human fixation collection procedure and the salient object labeling process. In the former procedure, the human fixation is captured when a viewer is displayed an image for 2-5 seconds under \textit{free-viewing} settings. Within such a short period of time, the viewer only fixates to some image locations that immediately attract his/her attention. For the latter process, a labeler is given a longer time to mark the pixels belonging to the salient object(s). In case of multiple objects appearing in the image, the labeler naturally identifies the \textit{semantic label} of each object and then decides which object is salient. This bridges the problem of salient object detection into the semantic segmentation research. In the latter semantic segmentation problem, the semantic label of each single pixel is decided based on a trained model which maps the features of the region containing the pixel and a particular semantic class label~\cite{LiuCe}. There are many improvements in this task by handling the adaptive inference~\cite{Adaptive}, adding object detectors~\cite{TigheCVPR2013}, or adopting deep superpixel's features~\cite{ICARCV}. There emerges a deep learning method, fully connected network (FCN)~\cite{Long}, which modifies the popular Convolutional Neural Networks (CNN)~\cite{CNN} to a new deep model mapping the input pixel directly to a semantic class label. There are many works improving FCN by considering more factors such as  probabilistic graphical models ~\cite{FCNCRF}.

Recently, along with the advancement of deep learning in semantic segmentation, deep networks, such as CNN, or even FCN, have been exploited to obtain more robust features than handcrafted ones for salient object detection. In particular, \textbf{deep networks}~\cite{LEGS,HKUIS,MTDS} achieve substantially better results than previous state of the art. However, these works only focus on switching the training data (with output from semantic classes to binary classes for salient object detection problem), or adding more network layers. In fact, the impact of the semantic information is not explicitly studied in the previous deep network-based saliency models. Therefore, in this work, we investigate the application of semantic information into the problem of salient object detection. In particular, we propose the \textit{semantic priors} to form the explicit and implicit semantic saliency maps in order to produce a high quality salient object detector. The main contributions of this work can be summarized as follows.

\begin{itemize}
\item We conduct the comprehensive study on how the semantic priors affect the salient object detection.
\item We propose the explicit saliency map and the implicit saliency map, derived from the semantic priors, which can discover the saliency object(s).
\item We extensively evaluate our proposed method on two challenging datasets in order to know the impact of our work in different settings.
\end{itemize}

\section{Proposed Method}
In this section, we describe the details of our proposed \textit{semantic priors} (SP) based saliency detection, and we show how to integrate the semantic priors as well as the saliency fusion can be efficiently computed. Figure \ref{fig:framework} illustrates the overview of our processing steps.

\subsection{Semantic Extraction}
\label{sec:21}
Saliency detection and semantic segmentation are highly correlated but essentially different in that saliency detection aims at separating generic salient objects from background, whereas semantic segmentation focuses on distinguishing objects of different categories. As aforementioned, fully connected network (FCN)~\cite{Long} and its variant, \textit{i.e.}, ~\cite{FCNCRF} are currently the state-of-the-art methods in the semantic segmentation task. Therefore, in this paper, we consider the end-to-end deep fully connected networks into our framework. Here, ``end-to-end'' means that the deep networks only need to be run on the input image once to produce a complete semantic map $C$ with the same pixel resolution as the input image.  We combine outputs from the final layer and the pool4 layer, at stride 16 and pool3, at stride 8. In particular, we obtain the confidence score $C_{x,y}$ for each single pixel $(x, y)$ as below.

\begin{equation}
C_{x,y} = \{C_{x,y}^1, C_{x,y}^2, \cdots, C_{x,y}^{n_c}\},
\end{equation}
where $C_{x,y}^1, C_{x,y}^2, \cdots, C_{x,y}^{n_c}$ indicate the likelihood that the pixel $(x,y)$ belongs to the listed $n_c$ semantic classes. Given an input image with size $h \times w$, the dimension of $C$ is $h \times w \times n_c$.

\subsection{Explicit Saliency Map}
\label{sec:22}
The objective of the explicit saliency map is to capture the preference of humans over different semantic classes. In other words, we aim to investigate which class is favoured by humans if there exist two or more classes in the input image. The class label $L_{x,y}$ of each single pixel $(x,y)$ can be obtained as below:

\begin{equation}
L_{x,y} = \arg\max C_{x,y}.
\end{equation}

Next, given a groundtruth map $G$, the density of each semantic class $k$ in the input image is defined by:

\begin{equation}
p_k = \frac{\sum_{x,y}(L_{x,y} = k) \times G_{x,y}}{\sum_{x,y}(L_{x,y} = k)}, 
\end{equation}
where $(L_{x,y} = k)$ is a boolean expression which verifies whether $L_{x,y}$ is equivalent to class $k$. Note that the size of the groundtruth map is also $h \times w$. Given the training dataset, we extract the co-occurrence saliency pairwise of one class and other $n_c - 1$ classes. The pairwise value $\theta_{g,t}$ of two semantic classes $g$ and $t$ is computed as below.

\begin{equation}
\theta_{k,t} =
  \begin{cases}
    1       & \quad , 	\exists L_{x',y'} = k \land L_{x'',y''} = t\\
    0  & \quad , \text{otherwise }\\
  \end{cases}.
\end{equation}

We define the \textit{explicit semantic priors} as the accumulated co-occurrence saliency pairwise of all classes. The explicit semantic priors of two classes $g$ and $t$ is calculated as below.

\begin{equation}
sp_{k,t}^{Explicit} = \frac{\sum_{i=1}^{n_t} p^i_k\theta^i_{k,t}}{\sum_{i=1}^{n_t}\theta^i_{k,t} + \epsilon},
\end{equation}
where $n_t$ is the number of images in the training set, and $\epsilon$ is inserted to avoid the division by zero. For the testing phase, given a test image, the explicit saliency value of each single pixel $(x,y)$ is computed as:
\begin{equation}
S^{Explicit}_{x,y} = \sum_{k = 1}^{n_c} \sum_{t = 1}^{n_c} (L_{x,y} = k) \times \theta_{k,t} \times sp^{Explicit}_{k,t}.
\end{equation}

\subsection{Implicit Saliency Map}
\label{sec:23}
Obviously the explicit saliency map performs well in case of the detected objects are in the listed class labels. However, the explicit saliency map fails in case of the salient objects are not in the $n_c$ class labels. Therefore, we propose the implicit saliency map which can uncover the salient objects not in the listed semantic classes. To this end, we oversegment the input image into non-overlapping regions. Then we extract features from each image region. Different from other methods which simply learn the mapping between the locally regional features with the saliency values, here, we take the semantic information into consideration. In particular, we are interested in studying the relationship between the regional features with the saliency values under the impact of semantic-driven features. Therefore, besides the off-the-shelf region features, we add two new features for each image region, namely, global semantic and local semantic features. The local semantic feature of each image region $q$ is defined as:

\begin{equation}
sp_{1,q} = \frac{\sum_{x,y}G_{x,y}\times (r(x,y) = q)}{\sum_{x,y} r(x,y) = q},
\end{equation}
where $r(x,y)$ returns the region index of pixel $(x,y)$. Meanwhile, the global semantic feature of the image region $q$ is defined as:

\begin{equation}
sp_{2,q} = \frac{\sum_{x,y}C_{x,y} \times (r(x,y) = q)}{h \times w}.
\end{equation}

The semantic features $sp^{Implicit} = \{sp_{1},sp_{2}\}$ are finally combined with other regional features. We consider the semantic features here as the \textit{implicit semantic priors} since they implicitly affect the mapping of the regional features and saliency scores. All of regional features are listed in Table~\ref{table:features}. Then, we learn a regressor $rf$ which maps the extracted features to the saliency values. In this work, we adopt the random forest regressor in~\cite{DRFI} which demonstrates a good performance. The training examples include a set of $n_r$ regions $\{\{{r_1},sp_1^{Implicit}\},\{{r_2},sp_2^{Implicit}\}, \cdots ,\{{r_{n_r}},sp_{n_r}^{Implicit}\}\}$ and the corresponding saliency scores $\{s_1,s_2, \cdots , s_{n_r}\}$, which are collected from the oversegmentation across a set of images with the ground truth annotation of the salient objects. The saliency value of each training image region is set as follows: if the number of the pixels (in the region) belonging to the salient object or the background exceeds 80\% of the number of the pixels in the region, its saliency value is set as 1 or 0, respectively.

\begin{table}[!t]
\caption{The regional features. Two sets of semantic features are included, namely $sp_1$ and $sp_2$. }

\begin{tabular}{|l|c|}
\hline
\textbf{Description}~~~~~~~~~~~~~~~~~~~~~~~~~~~~~~~~~~~~~~~~~~~~~~~~~~~~   & ~~\textbf{Dim}~~ \\ \hline \hline
The average normalized coordinates & 2    \\ \hline
The bounding box location & 4    \\ \hline
The aspect ratio of the bounding box & 1    \\ \hline
The normalized perimeter & 1    \\ \hline
The normalized area & 1    \\ \hline 
The normalized area of the neighbor regions & 1    \\ \hline
The variances of the RGB values & 3    \\ \hline
The variances of the L*a*b* values & 3    \\ \hline
The aspect ratio of the bounding box & 3    \\ \hline
The variances of the HSV values & 3    \\ \hline
Textons~\cite{Textons} & 15    \\ \hline
The local semantic features $sp_1$& $n_c$    \\ \hline
The global semantic features $sp_2$& $n_c$    \\ \hline
\end{tabular}
\label{table:features}
\end{table}

For the testing phase, given the input image, the implicit saliency value of each image region $q$ is computed by feeding the extracted features into the trained regressor $rf$:
\begin{equation}
S_q^{Implicit} = rf(\{{r_q},sp_q^{Implicit}\}).
\end{equation}

\begin{figure}[!t]
\centering
\includegraphics[width = \linewidth]{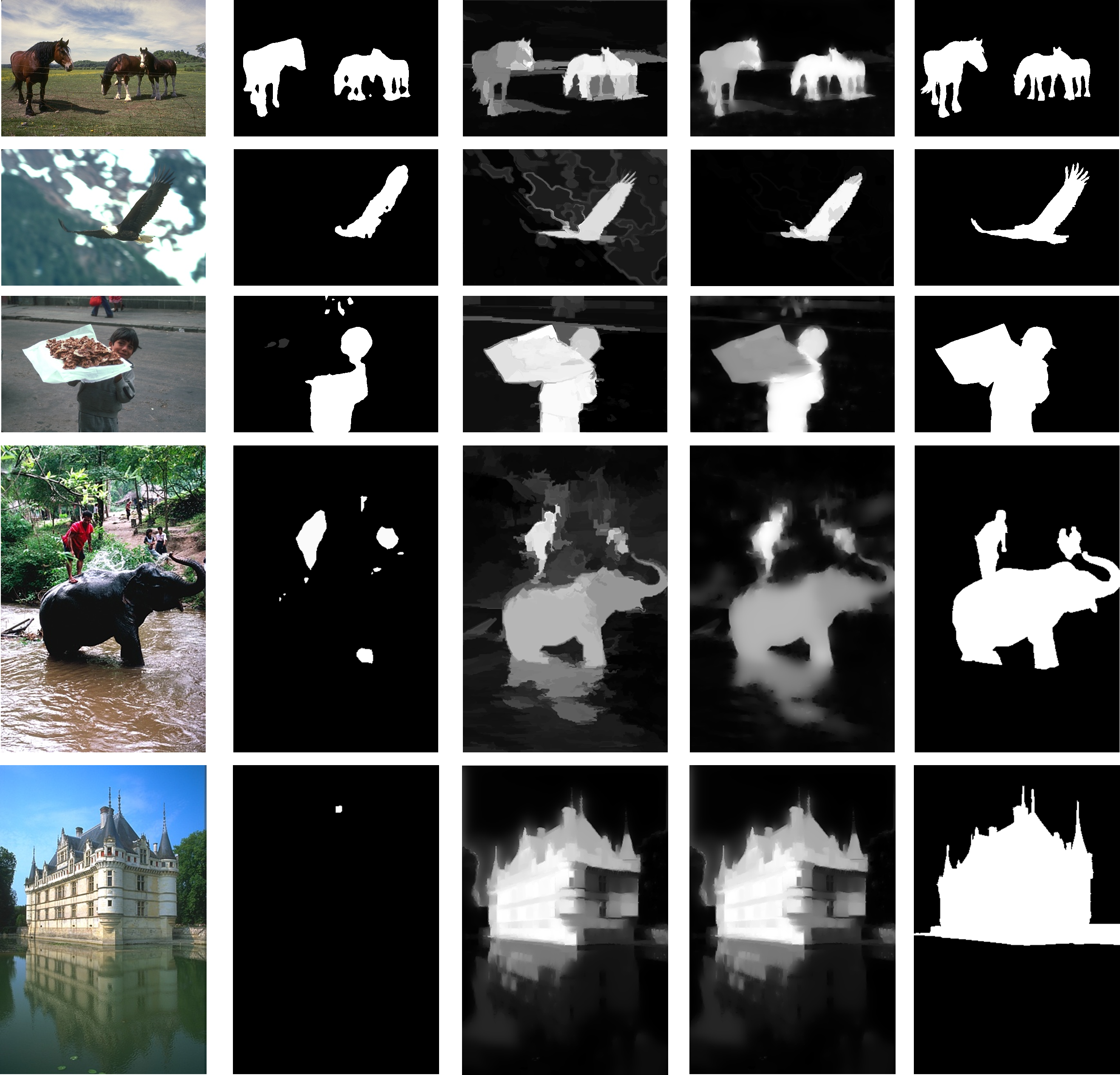}
%\vspace{-6mm}
\caption{From left to right: the original image, the explicit saliency map, the implicit saliency map, our final saliency map, the groundtruth map. From top to bottom: in the first two rows, the explicit map helps remove the background noise from the implicit map; (third row) the implicit map recovers the food tray held by the boy; (fourth row) the elephant is revealed owing to the implicit map; (fifth row) the building is fully recovered by the implicit map. Note that the \textit{food tray}, \textit{elephant}, and \textit{building} are not in the listed semantic classes of the PASCAL VOC dataset.}
\label{fig:extension}
\end{figure}

\begin{figure*}[!t]
\centering
\includegraphics[width = 0.975\linewidth]{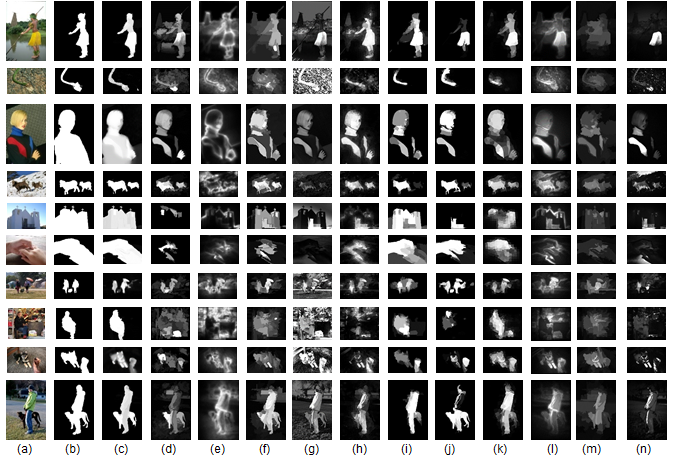}
\caption{Visual comparison of saliency maps. From left to right: (a) Original images, (b) ground truth, (c) our SP method, (d) BMS, (e) CA, (f) DRFI, (g) FT , (h) HDCT, (i) LEGS, (j) MDF, (k) MTDS, (l) PCA, (m) RC, (n) SF. Most results are of low resolution or highlight edges whereas our final result focuses on the main salient object as shown in ground truth map (c).}
\label{fig:results}
\end{figure*}

\subsection{Saliency Fusion}
\label{sec:24}
Given an input image with a size $h\times w$, the saliency maps from both aforementioned saliency maps are fused at the end. In particular, we scale the implicit saliency map $S^{Implicit}$, explicit saliency map $S^{Explicit}$, to the range [0..1]. Then we combine these maps as follows to compute a saliency value $S$ for each pixel:

\begin{equation}
S = \alpha S^{Explicit} + \gamma S^{Implicit}, 
\end{equation}
where the weights $\alpha$ is adaptively set as $\frac{\sum_{x,y}S^{Implicit}_{x,y}}{h \times w}$. Actually $\alpha$ measures how large the semantic pixels occupied in the image. Meanwhile, $\gamma$ is set as $1 - \alpha$. The resulting pixel-level saliency map may have an arbitrary scale. Therefore, in the final step, we rescale the saliency map to the range [0..1] or to contain at least 10\% saliency pixels. Fig.~\ref{fig:extension} demonstrates that the two individual saliency maps, \textit{i.e.}, explicit and implicit ones, complement each other in order to yield the good result.

\subsection{Implementation Settings}
\label{sec:25} 
For the implementation, we adopt the extension of FCN, namely CRF-FCN~\cite{FCNCRF}, to perform the semantic segmentation for the input image. In particular, we utilize the CRF-FCN model trained from the PASCAL VOC 2007 dataset~\cite{VOC} with 20 semantic classes\footnote{There are 20 semantic classes in the PASCAL VOC 2007 (`aeroplane', `bicycle', `bird', `boat', `bottle', `bus', `car', `cat', `chair', `cow', `diningtable', `dog', `horse', `motorbike', `person', `pottedplant', `sheep', `sofa', `train', `tvmonitor'); and an extra `others' class label.}. Therefore, the regional feature's dimensionality is $79$. We trained our SP framework on HKUIS dataset~\cite{HKUIS} (training part) which contains $4,000$ pairs of images and groundtruth maps. For the image over-segmentation, we adopt SLIC method~\cite{SLIC}. We set the number of regions as $200$ as a trade-off between the fine over-segmentation and the processing time. 

\section{Evaluation}
\subsection{Datasets and Evaluation Metrics}
We evaluate and compare the performances of our algorithm against previous baseline algorithms on two challenging benchmark datasets: ECSSD~\cite{HS} and HKUIS~\cite{HKUIS} (testing part). The ECSSD dataset contains 1,000 images with the complex and cluttered background. Meanwhile, the HKUIS contains 1,447 images. Note that each image in both datasets contains single or multiple salient objects.

The first evaluation compares the precision and recall rates. In the first setting, we compare binary masks for every threshold in the range [0..255]. In the second setting, we use the image dependent adaptive threshold proposed by~\cite{FT}, defined as twice the mean value of the saliency map $S$. In addition to precision and recall we compute their weighted harmonic mean measure or $F-measure$,
which is defined as:
\begin{equation}
F_{\beta} = \frac{(1 + \beta^2)\times Precision \times Recall}{\beta^2 \times Precision + Recall}.
\end{equation}
As in previous methods \cite{FT,SF}, we use $\beta^2$ = $0.3$.

For the second evaluation, we follow \cite{SF} to evaluate the mean absolute error (MAE) between a continuous saliency map $S$ and the binary ground truth $G$
for all image pixels $(x,y)$, defined as:
\begin{equation}
MAE = \frac{1}{h \times w}\sum_{x,y}{|S_{x,y} - G_{x,y}|}.
\end{equation}

\subsection{Performance on ECSSD dataset}

\begin{figure*}[!t]
\centering
\subfigure[Fixed threshold]{
\includegraphics[width = 0.31\linewidth]{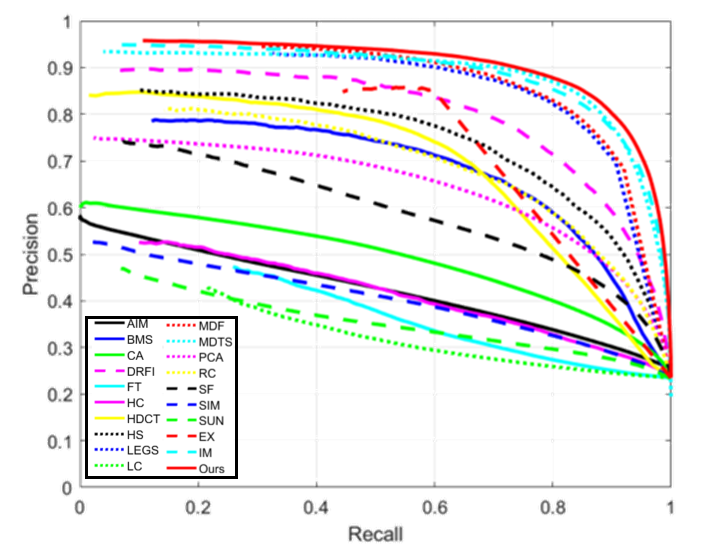}
	}
\subfigure[Adaptive threshold]{
\includegraphics[width = 0.33\linewidth]{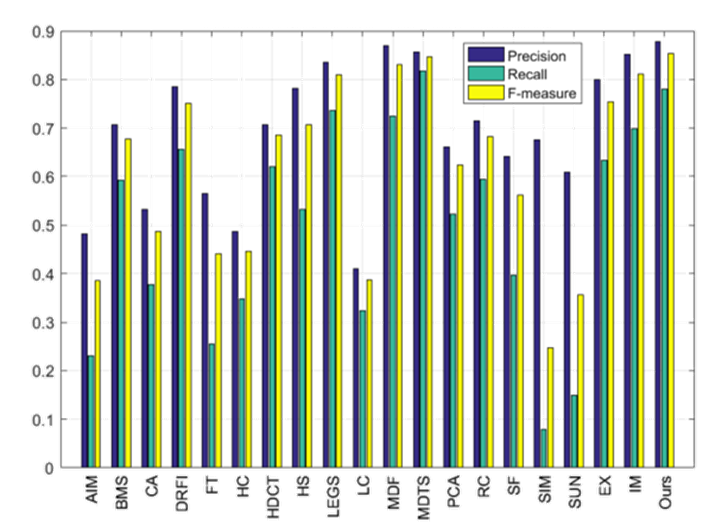}
}
\subfigure[Mean absolute error]{
\includegraphics[width = 0.315\linewidth]{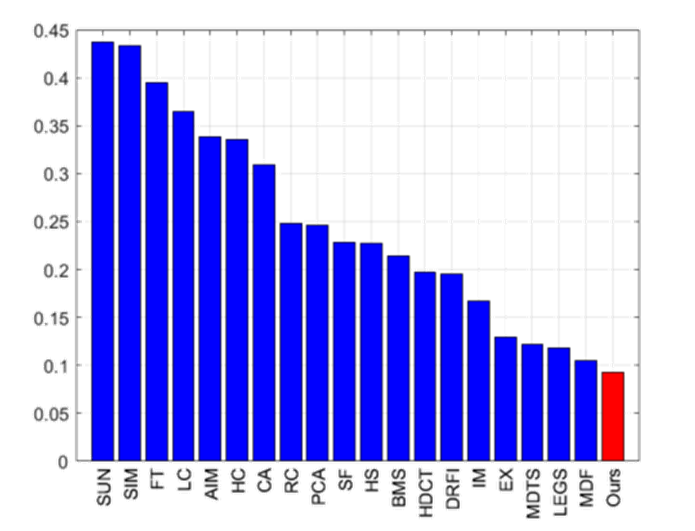}
}
\vspace{-3mm}
\caption{Statistical comparison with 17 saliency detection methods using all the $1,000$ images from ECSSD dataset \protect\cite{HS} with pixel accuracy saliency region annotation: (a) the average precision recall curve by segmenting saliency maps using fixed thresholds, (b) the average precision recall by adaptive thresholding (using the same method as in FT~\protect\cite{FT}, SF~\protect\cite{SF}, etc.), (c) the mean absolute error of the different saliency methods to ground truth mask. }
\label{fig:res_MSRA}
\end{figure*}

Following~\cite{HS}, we first evaluate our methods using a precision/recall curve which is shown in Figure~\ref{fig:res_MSRA}. We compare our work with $17$ state-of-the-art methods by running the approaches' publicly available source code: attention based on information maximization (AIM~\protect\cite{AIM}), boolean map saliency (BMS~\cite{BMS}), context-aware saliency (CA~\protect\cite{CA}), discriminative regional feature integration (DRFI~\cite{DRFI}), frequency-tuned saliency (FT~\protect\cite{FT}),global contrast saliency (HC and RC~\protect\cite{RC}), high-dimensional color transform (HDCT~\cite{HDCT}), hierarchical saliency (HS~\cite{HS}), spatial temporal cues (LC~\protect\cite{LC}),  local estimation and global search (LEGS~\cite{LEGS}), multiscale deep features (MDF~\cite{HKUIS}), multi-task deep saliency (MTDS~\cite{MTDS}), principal component analysis (PCA~\cite{PCA}), saliency filters (SF~\protect\cite{SF}), induction model (SIM~\protect\cite{SIM}), saliency using natural statistics (SUN~\protect\cite{SUN}. Note that LEGS, MDF, and MTDS are deep learning based methods. The visual comparison of saliency maps generated from our method and different baselines are demonstrated in Figure \ref{fig:results}. Our results are close to ground truth and focus on the main salient objects. As shown in Figure \ref{fig:res_MSRA}a, our work reaches the highest precision/recall rate over all baselines. As a result, our method also obtains the best performance in terms of F-measure. 

As discussed in the SF~\cite{SF}, neither the precision nor recall measure considers the true negative counts. These measures favor methods which successfully assign saliency to salient pixels but fail to detect non-salient regions over methods that successfully do the opposite. Instead, they suggested that MAE is a better metric than precision recall analysis for this problem. As shown in Figure \ref{fig:res_MSRA}c, our work outperforms the state-of-the-art performance \cite{HKUIS} by 10\%.

\begin{figure*}[!t]
\centering
\subfigure[Fixed threshold]{
\includegraphics[width = 0.32\linewidth]{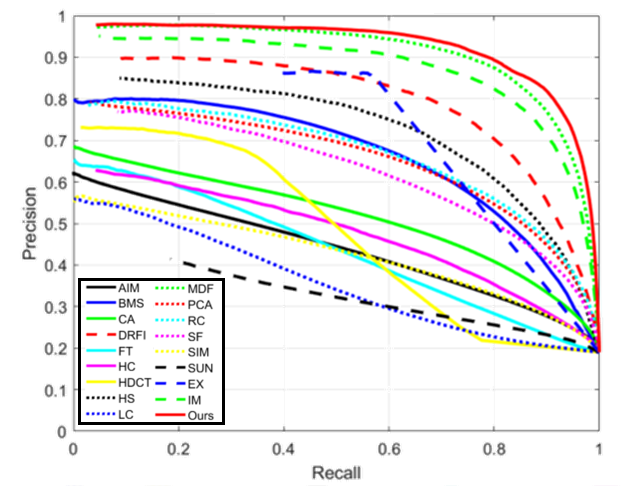}
}
\subfigure[Adaptive threshold]{
\includegraphics[width = 0.32\linewidth]{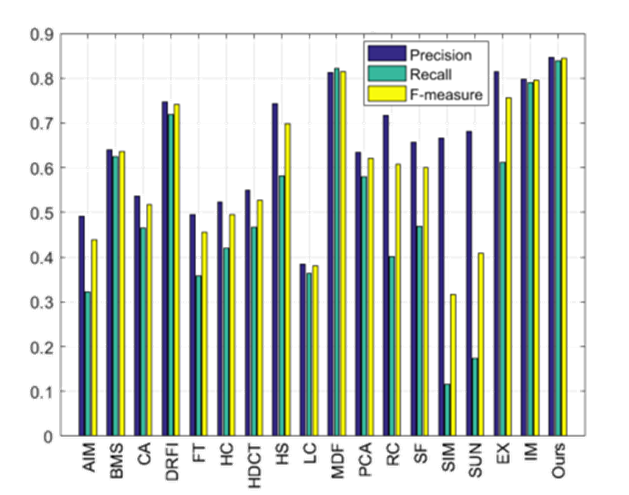}
}
\subfigure[Mean absolute error]{
\includegraphics[width = 0.32\linewidth]{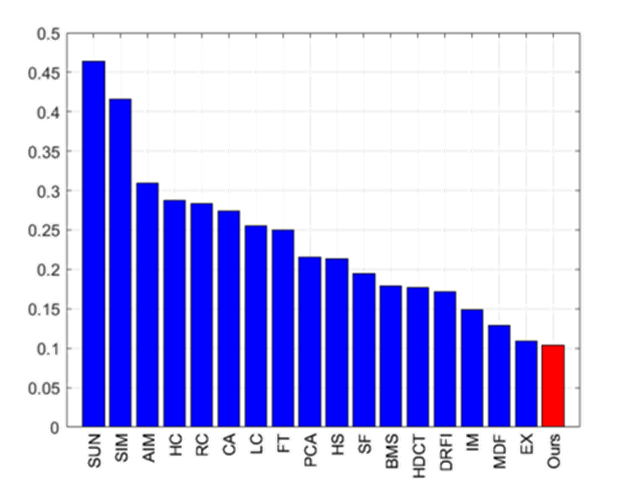}
}
\vspace{-3mm}
\caption{Statistical comparison with 15 saliency detection methods using all the $1,447$ images from the test set of HKUIS benchmark \protect\cite{HKUIS} with pixel accuracy saliency region annotation: (a) the average precision recall curve by segmenting saliency maps using fixed thresholds, (b) the average precision recall by adaptive thresholding (using the same method as in FT~\protect\cite{FT}, etc.), (c) the mean absolute error of the different saliency methods to ground truth mask.}
\label{fig:res_Icoseg}
\end{figure*}

\subsection{Performance on HKUIS dataset}
Since HKUIS is a relatively new dataset, we only have 15 baselines. We first evaluate our methods using a precision/recall curve which is shown in Figure~\ref{fig:res_Icoseg}a, b. Our method outperforms all other baselines in both two settings, namely fixed threshold and adaptive threshold. As shown in Figure \ref{fig:res_Icoseg}c, our method achieves the best performance in terms of MAE. In particular, our work outperforms other methods by a large margin, 25\%. 

\subsection{Effectiveness of Explicit and Implicit Saliency Maps}

We also evaluate the individual components in our system, namely, the explicit saliency map (EX), and the implicit saliency map (IM), in both ECSSD and HKUIS. As shown in Fig.~\ref{fig:res_MSRA} and Fig.~\ref{fig:res_Icoseg}, the two components generally achieve the acceptable performance (in terms of precision, recall, F-measure and MAE) which is comparable to other baselines. EX outperforms IM in terms of MAE, whereas IM achieves a better performance in terms of F-measure. When adaptively fusing them together, our unified framework achieves the state-of-the-art performance in every single evaluation metric. That demonstrates that these individual components complement each other in order to yield the good result.

\subsection{Computational Efficiency}
It is also worth investigating the computational efficiency of different methods. In Table \ref{table:times}, we compare the average running time for a typical $300 \times 400$ image of our approach to other methods. The average time is taken on a PC with Intel i7 $2.6$ GHz CPU and $8$GB RAM with our unoptimized Matlab code. Performance of all the methods compared in this table are based on implementations in C++ and Matlab. Basically, C++ implementation runs faster than the Matlab based code. The CA method~\cite{CA} is the slowest one because it requires an exhaustive nearest-neighbor search among patches. Meanwhile, our method is able to run faster than other Matlab based implementations. Our procedure spends most of the computation time on semantic segmentation and extracting regional features.

\section{Conclusion and Future Work}
In this paper, we have presented a novel method, \textit{semantic priors} (SP), which adopts the semantic segmentation in order to detect \textit{salient objects}. To this end, two maps are derived from semantic priors: the explicit saliency map and the implicit saliency map. These two maps are fused together to give a saliency map of the salient objects with sharp boundaries.  Experimental results on two challenging datasets demonstrate that our salient object detection results are 10\% - 25\% better than the previous best results (compared against 15+ methods in two different datasets), in terms of mean absolute error. 

For future work, we aim to investigate other sophisticated techniques for semantic segmentation with a larger number of semantic classes. Also, we would like to study the reverse impact of salient object detection into the semantic segmentation process.  

\begin{table}[!t]
\caption{Runtime comparison of different methods.}
\begin{tabular}{|l|c|c|c|c|c|}
\hline
Method~~   & ~~CA~~     & ~~DRFI~~   & ~~SF~~   & ~~RC~~   & ~~Ours~~ \\ \hline
Time (s)~~ & 51.2   & 10.0 & 0.15 & 0.25 & 3.8 \\ \hline
Code     & Matlab & Matlab  & C++  & C++  & Matlab  \\ \hline
\end{tabular}
\label{table:times}
\end{table}

\section*{Additional Authors}
Thanh Duc Ngo, Khang Nguyen from Multimedia Communications Laboratory, University of Information Technology are additionally supportive  authors. We gratefully acknowledge the support of NVIDIA Corporation with the GPU donation.
\small
\bibliographystyle{named}
\bibliography{egbib}
\linespread{1}
\end{document}